\pdfoutput=1

\documentclass[11pt]{article}

\usepackage{EMNLP2023}

\usepackage{times}
\usepackage{latexsym}

\usepackage[T1]{fontenc}

\usepackage[utf8]{inputenc}

\usepackage{microtype}

\usepackage{inconsolata}

\usepackage{enumitem}
\usepackage{pgfplots}
\usepackage{tabularx}
\usepackage{caption}
\usepackage{subcaption}
\usepackage{bbm}
\usepackage{url}
\usepackage{amsmath} 
\usepackage{pifont}
\usepackage{arydshln}
\usepackage{makecell}
\usepackage{booktabs}
\usepackage{multirow} 
\title{Self-Consistency Boosts Calibration for Math Reasoning}

\author{Ante Wang$^{1,2}$, Linfeng Song$^{3}$, Ye Tian$^{3}$, Baolin Peng$^{3}$, Lifeng Jin$^{3}$, Haitao Mi$^{3}$,\\
{\bf Jinsong Su}$^{1,2}$ and {\bf Dong Yu}$^{3}$ \\
$^{1}$School of Informatics, Xiamen University, China \\
$^{2}$Key Laboratory of Digital Protection and Intelligent Processing of Intangible Cultural Heritage\\of Fujian and Taiwan (Xiamen University), Ministry of Culture and Tourism, China \\
$^{3}$Tencent AI Lab, Bellevue, WA \\
\texttt{wangante@stu.xmu.edu.cn, lfsong@global.tencent.com} \\}

\begin{document}
\maketitle
\begin{abstract}
Calibration, which establishes the correlation between accuracy and model confidence, is important for LLM development.
We design three off-the-shelf calibration methods based on self-consistency \cite{wang2022self} for math reasoning tasks.
Evaluation on two popular benchmarks (GSM8K and MathQA) using strong open-source LLMs (Mistral and LLaMA2), our methods better bridge model confidence and accuracy than existing methods based on \emph{p(True)} \cite{kadavath2022language} or \emph{logit} \cite{guo2017calibration}.
\end{abstract}

\section{Introduction}

Mathematical reasoning tasks \cite{cobbe2021training,hendrycks2021measuring,amini2019mathqa} involve mapping a question into a series of equations, which are then solved to obtain the final answer.
Math reasoning has long been recognized challenging.
Existing solutions propose to map input questions into equations via semantic parsing \cite{matsuzaki2017semantic,hopkins2017beyond} or AST decoding \cite{li2019modeling,qin2021neural,wu2021math}.
Yet, the performance can degradate dramatically even with slight changes to the questions \cite{patel2021nlp,li2022seeking}.

Recently, large language models (LLM, \citealt{achiam2023gpt,touvron2023llama,jiang2024mixtral}) have shown great potential for solving many math reasoning tasks, even though they are not specifically trained on these tasks.
For instance, with chain-of-thought prompting \cite{wei2022chain} and self-consistency \cite{wang2022self}, open-source LLMs, such as Mixtral 8$\times$7B \cite{jiang2024mixtral}, can reach an accuracy of around 80\% on the GSM8K benchmark \cite{cobbe2021training}.
On the other hand, conventional pretrained models (e.g., T5 \cite{raffel2020exploring}) that are specifically finetuned on the GSM8K training set can only report accuracies around 10\% to 20\% \cite{shridhar2023distilling,magister2023teaching}.

However, LLMs lack adequate calibration out of the box -- the probabilities of model predictions are often poorly aligned with the actual accuracy \cite{xiong2023can,chen2023close}.
Calibration is important for LLM development, as a well-calibrated LLM can precisely \emph{tell} how likely its responses are correct or not.
With such information, LLM developers can take multiple options to handle low-confidence responses, such as letting the LLM refuse to answer or keep resampling until a confident response is produced.

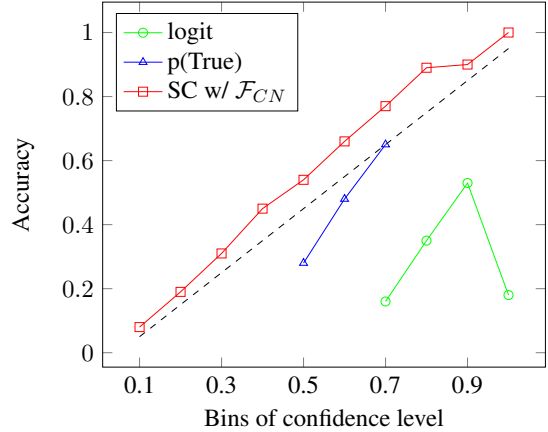
\begin{figure}
\begin{tikzpicture}[scale=0.85]
\begin{axis}[
    xlabel={Bins of confidence level},
    ylabel={Accuracy},
    legend pos=north west,
    legend style={cells={anchor=west}},
    xticklabels={,0.1,0.3,0.5,0.7,0.9}
]
\addplot[
    color=green,
    mark=o,
    ]
    coordinates {
        (6, 0.16)
        (7, 0.35)
        (8, 0.53)
        (9, 0.18)
    };
    \addlegendentry{logit} 

\addplot[
    color=blue,
    mark=triangle,
    ]
    coordinates {
        (4, 0.28)
        (5, 0.48)
        (6, 0.65)
    };
    \addlegendentry{p(True)}

\addplot[
    color=red,
    mark=square,
    ]
    coordinates {
        (0, 0.08)
        (1, 0.19)
        (2, 0.31)
        (3, 0.45)
        (4, 0.54)
        (5, 0.66)
        (6, 0.77)
        (7, 0.89)
        (8, 0.9)
        (9, 1.0)
    };
    \addlegendentry{SC w/ $\mathcal{F}_{CN}$}    

\addplot[
    dashed,
    color=black,
    ]
    coordinates {
        (0, 0.05)
        (1, 0.15)
        (2, 0.25)
        (3, 0.35)
        (4, 0.45)
        (5, 0.55)
        (6, 0.65)
        (7, 0.75)
        (8, 0.85)
        (9, 0.95)
    };

\end{axis}
\end{tikzpicture}
\caption{Comparison of several calibration methods on Mistral-7B, where \emph{SC w/ $\mathcal{F}_{CN}$} is one of our methods based on self-consistency, which will be introduced in \S \ref{sec:method}.}
\label{fig:three_lines}
\end{figure}

In this work, we propose calibration methods based on self-consistency \cite{wang2022self} for math reasoning tasks.
Self-consistency performs clustering over multiple LLM samples before picking one from the largest cluster as the response to each input query.
Here we consider several ways to estimate model confidence using the clustering results: \emph{cluster size} that estimates how many samples agree with the selected one, \emph{cluster number} that measures to what extent samples disagree with each other, and \emph{pairwise comparison} that captures relative differences between pairs of clusters.

We conduct experiments using strong open-source LLMs: 
Mistral \cite{jiang2023mistral,jiang2024mixtral} and LLaMA2 \cite{touvron2023llama} series models with\,/\,without being aligned with instructions.
Results on GSM8K \cite{cobbe2021training} and MathQA \cite{amini2019mathqa} show that all our methods better calibrate these models than exiting popular methods, such as \emph{p(True)} \cite{kadavath2022language} and \emph{logit} \cite{guo2017calibration} over the whole reasoning path or target answer span only.

\section{Preview: Self-Consistency with CoT Prompting}
For math reasoning, there are usually multiple trajectories to reach the final solution.
To replicate this process, \citet{wang2022self} initially sample various reasoning paths $r_1,...,r_N$ from the LLM given input $x$ with Chain-of-Thought (CoT) prompting.\footnote{Here we follow common practice to adopt demonstrations with rationales for pretrained only models (e.g., Mistral-7B) and use ``\emph{Let's think step by step}'' \cite{kojima2022large} for instruction-tuned models (e.g., Mistral-7B-Inst).}
Then, the answers $a_1,...,a_N$ are extracted from the paths,
and the most consistent answer (the one win by majority vote among the answers) is selected as the final answer $\mathbf{a}$:
\begin{equation}
\begin{aligned}
    \mathbf{a} = \max_{\hat{a}} \sum_{i=1}^{N} \mathbbm{1}( a_i = \hat{a}), \\
    r_i, a_i \sim \mathtt{LLM}_{\theta}(x),
\end{aligned}
\end{equation}
where $r_i$, $a_i$ denote the $i$-th sampled reasoning path and its corresponding answer, respectively.

\section{Calibration using Self-Consistency}
\label{sec:method}
After performing self-consistency on input $x$ using $\mathtt{LLM}_{\theta}$, we obtain a set of clusters $\mathcal{C} = \{c_1,..., c_{|\mathcal{C}|}\}$ with each cluster $c_i$ comprising $n_i$ sampled responses with the same answers.
We design the following strategies, tailored to the characteristics of these clusters, to estimate the confidence of $\mathtt{LLM}_{\theta}$.

\paragraph{Cluster Number}
We initially consider the \emph{Cluster Number} $|\mathcal{C}|$.
This is motivated by the finding of previous work \cite{wang2022self,xiong2023can}: LLMs tend to generate consistent answers when they are confident about their predictions, and thus the cluster number (number of distinct answers) tends to be small.
We further divide the cluster number by the sample size $N$ to normalize the score into the range of $[0, 1]$, before reversing it by ``$1-x$'':
\begin{equation}
\begin{aligned}
    \mathcal{F}_{CN}(x, \theta) = 1 - \frac{|\mathcal{C}|}{N}.
\end{aligned}
\end{equation}

\paragraph{Cluster Size}
In a similar vein, we adopt the \emph{Cluster Size}: the number of samples (e.g., $n_i$) within a specific cluster (e.g., $c_i$). 
Again, we compute its proportion relative to the total sample size to normalize the score into the range $[0, 1]$:
\begin{equation}
\begin{aligned}
    \mathcal{F}_{CS}(x, \theta) = \frac{n_i}{N}.
\end{aligned}
\end{equation}
In contrast to the cluster number, the cluster size is more universally applicable across diverse prompts, as the cluster number can easily become ineffective when the output space of an LLM is restricted, such as when options for a question are provided.

\paragraph{Pairwise Comparison}
The \emph{Cluster Number} and \emph{Cluster Size} primarily consider the number of distinct answers and the number of sampled paths within a single cluster, respectively.
They both overlook the information by comparing different clusters.
For example, they may fail to consider the situation when the sizes of the top-ranked clusters are close. Consequently, we introduce the \emph{Pairwise Comparison} method, which computes the winning rate of the chosen cluster ($c_i$) against each of the remaining clusters:
\begin{equation}
    \mathcal{F}_{PC}(x, \theta) = \prod_{j \neq i}^{|\mathcal{C}|} \frac{n_i}{n_i + n_j},
\end{equation}
where $\frac{n_i}{n_i + n_j}$ represents the winning rate of selected cluster $c_i$ against another cluster $c_j$.

\section{Experiments}

\begin{table*}[]
\small
\centering
\begin{tabular}{llcccccccc}
\toprule
& & \multicolumn{2}{c}{Mistral-7B} & \multicolumn{2}{c}{Mistral-7B-Inst} & \multicolumn{2}{c}{Mixtral-8$\times$7B} & \multicolumn{2}{c}{Mixtral-8$\times$7B-Inst} \\ 
\cmidrule(lr){3-4} \cmidrule(lr){5-6} \cmidrule(lr){7-8} \cmidrule(lr){9-10}
& & ECE $\downarrow$ & Brier $\downarrow$ & ECE $\downarrow$ & Brier $\downarrow$ & ECE $\downarrow$ & Brier $\downarrow$ & ECE $\downarrow$ & Brier $\downarrow$ \\
\midrule
\multirow{7}*{GSM8K}
& logit w/ Path & 0.394 & 0.399 & 0.414 & 0.414 & 0.178 & 0.265 & 0.233 & 0.252 \\
& logit w/ Answer & 0.505 & 0.488 & 0.467 & 0.458 & 0.307 & 0.312 & 0.236 & 0.238 \\
& p(True) & 0.127 & 0.267 & 0.406 & 0.407 & \textbf{0.070} & 0.201 & 0.195 & 0.198 \\
\cmidrule(lr){2-10}
& \emph{Self-Consistency} \\
& ~~~~w/ $\mathcal{F}_{CN}$ & \textbf{0.092} & 0.186 & \textbf{0.125} & 0.182 & 0.136 & 0.157 & \textbf{0.075} & 0.092 \\
& ~~~~w/ $\mathcal{F}_{CS}$ & 0.148 & \textbf{0.185} & 0.163 & \textbf{0.180} & 0.173 & \textbf{0.156} & 0.085 & \textbf{0.086} \\
& ~~~~w/ $\mathcal{F}_{PC}$ & 0.248 & 0.229 & 0.253 & 0.226 & 0.238 & 0.194 & 0.110 & 0.096 \\
\midrule
\midrule
\multirow{7}*{MathQA}
& logit w/ Path & 0.500 & 0.499 & 0.539 & 0.510 & 0.333 & 0.380 & 0.364 & 0.373 \\
& logit w/ Answer & 0.356 & 0.362 & 0.291 & 0.319 & 0.266 & 0.290 & 0.220 & 0.281 \\
& p(True) & 0.350 & 0.309 & 0.271 & 0.317 & 0.228 & 0.253 & 0.273 & 0.272 \\
\cmidrule(lr){2-10}
& \emph{Self-Consistency} \\
& ~~~~w/ $\mathcal{F}_{CN}$ & 0.331 & 0.336 & 0.374 & 0.359 & 0.143 & 0.236 & 0.128 & 0.215 \\
& ~~~~w/ $\mathcal{F}_{CS}$ & 0.091 & 0.225 & 0.114 & 0.227 & \textbf{0.080} & \textbf{0.190} & \textbf{0.035} & \textbf{0.171} \\
& ~~~~w/ $\mathcal{F}_{PC}$ & \textbf{0.052} & \textbf{0.220} & \textbf{0.065} & \textbf{0.219} & 0.143 & 0.203 & 0.054 & 0.174 \\
\bottomrule
\end{tabular}
\caption{Main test results on GSM8K and MathQA when using Mistral family models. Specifically, $*$-Inst indicates instruction-tuned models.}
\label{tab:main}
\end{table*}

\subsection{Setup}

\paragraph{Datasets}
We conduct experiments on two popular math reasoning benchmarks of different type of questions, GSM8K \cite{cobbe2021training} and MathQA \cite{amini2019mathqa}. 
Particularly, GSM8K comprises 1,319 linguistically diverse grade school math word problems for testing. 
On the other hand, MathQA offers 2,985 \emph{multiple-choice} math word problems for evaluation.

\paragraph{Evaluation Metrics}
We adopt Brier Score and Expected Calibration Error (ECE) as evaluating metrics following common practice \cite{geng2023survey}.

Given instances $(x_1, y_1),...,(x_{\bar{N}}, y_{\bar{N}})$ and their corresponding LLM predictions $\hat{y}_1,...,\hat{y}_{\bar{N}}$, ECE is computed by first binning the predictions into $M=10$ intervals based on their LLM confidence levels (e.g., $p(\hat{y}_i)$).
For each bin (e.g. $B_m$), it then calculates the accuracy (acc$(B_m)$) and the average confidence (conf$(B_m)$):
\begin{equation}
\begin{aligned}
\text{acc}(B_m) &= \frac{1}{|B_m|} \sum_{i \in B_m} \mathbbm{1}(y_i = \hat{y}_i), \\
\text{conf}(B_m) &= \frac{1}{|B_m|} \sum_{i \in B_m} p(\hat{y}_i),
\end{aligned}
\end{equation}
where $|B_m|$ is the number of samples in bin $B_m$.
Finally, the difference between accuracy and confidence is averaged across all bins to obtain the ECE score:
\begin{equation}
    \text{ECE} = \sum_{m=1}^{M} \frac{|B_m|}{\bar{N}} |\text{acc}(B_m) - \text{conf}(B_m)|
\end{equation}

As another popular metric, Brier score is similar to ECE but conducted at the instance level:
\begin{equation}
\text{Brier} = \frac{1}{\bar{N}} \sum_{i=1}^{\bar{N}} (p(\hat{y}_i) - \mathbbm{1}(y_i = \hat{y}_i))^2.
\end{equation}
Both metrics range from 0 to 1 with lower values indicating better calibration.
We take Brier score as the \emph{main} metric, as it is more robust to unbalanced distribution across bins (e.g. instances concentrate to one or two bins).

\paragraph{Settings}
We conduct experiments on LLaMA2 and Mistral-family models and investigate both pretrained or instruction-tuned variations.
We use nucleus sampling to obtain $N=16$ samples by default for each instance and use temperatures of 0.6 / 1.0 for all pretrained / instruction-tuned models.

\paragraph{Baselines}
We take the three representative baselines below for comparison:
\begin{itemize}[leftmargin=0.5cm]
\item \emph{logit w/ Path}: It averages the probabilities of the tokens from the whole path to estimate the confidence of each prediction.
\item \emph{logit w/ Answer}: It is similar to \emph{logit w/ Path} but only consider the tokens from the predicted answer span.
\item \emph{p(True)}: It asks the LLM itself to classify its prediction as \emph{True} or \emph{False}. Then, it takes the predicted probability of \emph{True} as its confidence. We follow \citet{kadavath2022language} to construct 8-shot demonstrations for prompting pretrained models but directly use instruction for instruction-tuned models.
\end{itemize}

\subsection{Results and Analysis}
\paragraph{Main Results}
Table \ref{tab:main} presents the main results obtained from both benchmarks using Mistral-family models. \emph{p(True)} performs best among the baselines, echoing the findings of \citet{kadavath2022language}. However, due to its reliance on prompt design and in-context examples to aid the LLM to classify its predictions, it can be challenging to construct effective demonstrations or instructions.

In general, self-consistency-based methods surpass baselines in most cases regarding Brier and ECE, validating the efficacy of employing self-consistency features for estimating model confidence. We also note that baselines can occasionally yield impressive ECE scores (\emph{p(True)} on GSM8K with Mixtral-8$\times$7B). However, we observe that this is attributed to the concentration of most samples in just a few bins (e.g., Figure \ref{fig:three_lines}), leading to unreliable measurements. Nevertheless, our approaches still exhibit strong performance in terms of ECE scores across various settings.

Among the self-consistency-based methods, $\mathcal{F}_{CN}$ yields better ECE results on GSM8K, while $\mathcal{F}_{CS}$ achieves the highest Brier score. 
Conversely, for MathQA, $\mathcal{F}_{CN}$ performs significantly worse than the other two.
This is because MathQA is a \emph{multi-choice} task, and thus the cluster number of LLM answers is strictly limited by the provided choices.
In conclusion, $\mathcal{F}_{CS}$ demonstrates greater generality across diverse settings, but $\mathcal{F}_{CN}$ and $\mathcal{F}_{PC}$ do offer improved estimation in certain cases.

\paragraph{Influence of Sample Size $N$}

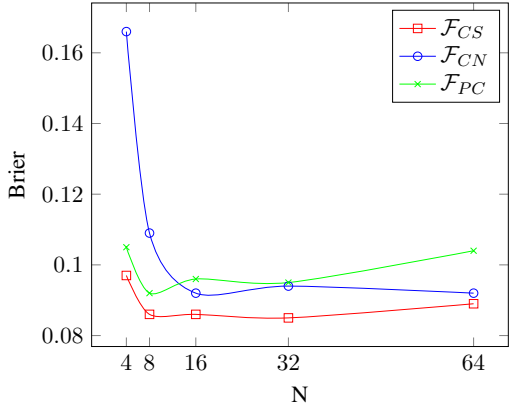
\begin{figure}
\pgfplotsset{
/pgf/number format/fixed,
/pgf/number format/precision=2
}
\begin{tikzpicture}[scale=0.8]
\begin{axis}[
xlabel=N,
ylabel=Brier,
xtick={4,8,16,32,64},
legend style={cells={anchor=west}},
]
\addplot[smooth,mark=square,red]
coordinates{
(4, 0.097)
(8, 0.086)
(16, 0.086)
(32, 0.085)
(64, 0.089)
}; \addlegendentry{$\mathcal{F}_{CS}$}
\addplot[smooth,mark=o,blue]
coordinates{
(4, 0.166)
(8, 0.109)
(16, 0.092)
(32, 0.094)
(64, 0.092)
}; \addlegendentry{$\mathcal{F}_{CN}$}
\addplot[smooth,mark=x,green]
coordinates{
(4, 0.105)
(8, 0.092)
(16, 0.096)
(32, 0.095)
(64, 0.104)
}; \addlegendentry{$\mathcal{F}_{PC}$}
\end{axis}
\end{tikzpicture}
\caption{Calibration results on GSM8K when using Mixtral-8$\times$7B-Inst with different $N$.}
\label{fig:Mixtral_N}
\end{figure}

Previous research \cite{wang2022self} has demonstrated that the sample size $N$ can significantly affect the accuracy of self-consistency. When $N$ increases, the model performance initially continues to improve before stabilizing once $N$ reaches a sufficient level. Therefore, we take Mixtral-8$\times$7B-Inst as a case study to examine the impact of $N$ on calibration.

As illustrated in Figure \ref{fig:Mixtral_N}, the Brier scores for all our methods initially decline and then remain constant as $N$ grows. For $\mathcal{F}_{CS}$ and $\mathcal{F}_{PC}$, $N=8$ is adequate for accurate estimation. In contrast, $\mathcal{F}_{CN}$ requires a larger $N$, indicating that the cluster number is more susceptible to the randomness of sampling.

\paragraph{Correlation between Performance and Calibration}

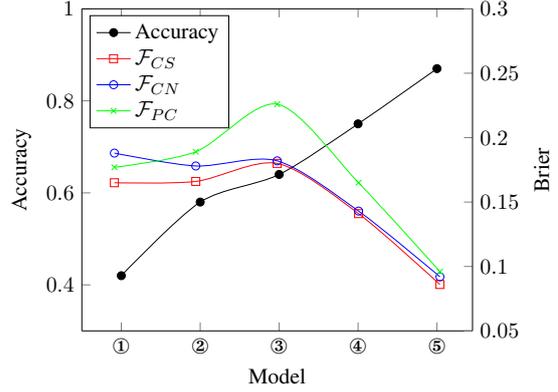
\begin{figure}
\centering
\pgfplotsset{
compat=1.3,
/pgf/number format/fixed,
/pgf/number format/precision=2
}
\begin{tikzpicture}[scale=0.75]
\begin{axis}[
axis y line*=left,
ylabel=Accuracy,
xlabel=Model,
xtick={0,...,5},
xmin=-0.5,
xticklabels={\ding{172},\ding{173},\ding{174},\ding{175},\ding{176},},
ymin=0.3, ymax=1.0
]
\addplot[smooth,mark=*,black]
coordinates{
(0, 0.42)
(1, 0.58)
(2, 0.64)
(3, 0.75)
(4, 0.87)
}; \label{plot_one}
\end{axis}

\begin{axis}[
axis y line*=right,
axis x line=none,
ylabel=Brier,
ymin=0.05, ymax=0.3,
legend style={at={(0.02,0.98)}, anchor=north west, cells={anchor=west}}
]
\addlegendimage{/pgfplots/refstyle=plot_one}\addlegendentry{Accuracy}
\addplot[smooth,mark=square,red]
coordinates{
(0, 0.165)
(1, 0.166)
(2, 0.180)
(3, 0.141)
(4, 0.086)
}; \addlegendentry{$\mathcal{F}_{CS}$}

\addplot[smooth,mark=o,blue]
coordinates{
(0, 0.188)
(1, 0.178)
(2, 0.182)
(3, 0.143)
(4, 0.092)
}; \addlegendentry{$\mathcal{F}_{CN}$}

\addplot[smooth,mark=x,green]
coordinates{
(0, 0.177)
(1, 0.189)
(2, 0.226)
(3, 0.165)
(4, 0.096)
}; \addlegendentry{$\mathcal{F}_{PC}$}
\end{axis}
\end{tikzpicture}
\caption{Performance and calibration results on GSM8K using different models below sorted by their performance: \ding{172} LLaMA2-7B-Chat, \ding{173} LLaMA2-13B-Chat, \ding{174} Mistral-7B-Inst, \ding{175} LLaMA2-70B-Chat, \ding{176} Mixtral-8$\times$7B-Inst.}
\label{fig:Acc_Brier}
\vspace{-1.0em}
\end{figure}

We finally explore the associations between model performance (Accuracy) and calibration. Figure \ref{fig:Acc_Brier} showcases the results on instruction-tuned LLaMA2 and Mistral series models, arranged in ascending order based on their performance.
We generally observe a positively correlated trend between calibration (lower the better) and performance (higher the better) among the studied models.
This observation indicates that more powerful models also exhibit enhanced calibration, echoing the findings of \citet{kadavath2022language}. This phenomenon can be attributed to the fact that when a tested LLM is stronger, it is capable of generating more reasonable and consistent responses, leading to improved calibration.

\section{Conclusion}
In this paper, we extend the widely-used inference strategy, self-consistency, to the field of calibration. Specifically, we develop three off-the-shelf calibration methods based on self-consistency for math reasoning tasks. Compared to conventional methods (\emph{p(True)} and \emph{logit}), our approaches yield significantly improved ECE and Brier scores on popular GSM8K and MathQA datasets. 
Future research directions include designing more effective calibration methods, leveraging richer features and employing more strategies (e.g., \emph{temperature scaling} \cite{guo2017calibration}) to enhance calibration performance. 
Our ultimate goal is to construct reliable and honest LLMs with the help of accurate confidence estimation.

\section*{Limitations}
Our methods are founded on the principle of self-consistency, which relies on sampling multiple times for prediction. This approach, however, needs additional cost for inference, which may not be efficient and eco-friendly. Besides, our current work is limited to mathematical problems and does not explore other types of tasks, such as question-answering. Although it is crucial to extend our methods to encompass other tasks, this is non-trivial due to the inherent difficulty in dividing certain tasks' model predictions into distinct clusters.

\section*{Ethics Statement}
We focus on ethical AI research and strive to achieve a balance between technological advancements and our ethical responsibilities.
This work studies calibration, which aims to enhance the reliability of LLMs.
Besides, we conduct experiments only on publicly available datasets, upholding privacy and anonymity rules.

\bibliography{custom}

\begin{thebibliography}{25}
\expandafter\ifx\csname natexlab\endcsname\relax\def\natexlab#1{#1}\fi

\bibitem[{Achiam et~al.(2023)Achiam, Adler, Agarwal, Ahmad, Akkaya, Aleman, Almeida, Altenschmidt, Altman, Anadkat et~al.}]{achiam2023gpt}
Josh Achiam, Steven Adler, Sandhini Agarwal, Lama Ahmad, Ilge Akkaya, Florencia~Leoni Aleman, Diogo Almeida, Janko Altenschmidt, Sam Altman, Shyamal Anadkat, et~al. 2023.
\newblock Gpt-4 technical report.
\newblock \emph{arXiv preprint arXiv:2303.08774}.

\bibitem[{Amini et~al.(2019)Amini, Gabriel, Lin, Koncel-Kedziorski, Choi, and Hajishirzi}]{amini2019mathqa}
Aida Amini, Saadia Gabriel, Shanchuan Lin, Rik Koncel-Kedziorski, Yejin Choi, and Hannaneh Hajishirzi. 2019.
\newblock Mathqa: Towards interpretable math word problem solving with operation-based formalisms.
\newblock In \emph{Proceedings of the 2019 Conference of the North American Chapter of the Association for Computational Linguistics: Human Language Technologies, Volume 1 (Long and Short Papers)}, pages 2357--2367.

\bibitem[{Chen et~al.(2023)Chen, Yuan, Cui, Liu, and Ji}]{chen2023close}
Yangyi Chen, Lifan Yuan, Ganqu Cui, Zhiyuan Liu, and Heng Ji. 2023.
\newblock A close look into the calibration of pre-trained language models.
\newblock In \emph{Proceedings of the 61st Annual Meeting of the Association for Computational Linguistics (Volume 1: Long Papers)}, pages 1343--1367.

\bibitem[{Cobbe et~al.(2021)Cobbe, Kosaraju, Bavarian, Chen, Jun, Kaiser, Plappert, Tworek, Hilton, Nakano et~al.}]{cobbe2021training}
Karl Cobbe, Vineet Kosaraju, Mohammad Bavarian, Mark Chen, Heewoo Jun, Lukasz Kaiser, Matthias Plappert, Jerry Tworek, Jacob Hilton, Reiichiro Nakano, et~al. 2021.
\newblock Training verifiers to solve math word problems.
\newblock \emph{arXiv preprint arXiv:2110.14168}.

\bibitem[{Geng et~al.(2023)Geng, Cai, Wang, Koeppl, Nakov, and Gurevych}]{geng2023survey}
Jiahui Geng, Fengyu Cai, Yuxia Wang, Heinz Koeppl, Preslav Nakov, and Iryna Gurevych. 2023.
\newblock A survey of language model confidence estimation and calibration.
\newblock \emph{arXiv preprint arXiv:2311.08298}.

\bibitem[{Guo et~al.(2017)Guo, Pleiss, Sun, and Weinberger}]{guo2017calibration}
Chuan Guo, Geoff Pleiss, Yu~Sun, and Kilian~Q Weinberger. 2017.
\newblock On calibration of modern neural networks.
\newblock In \emph{International conference on machine learning}, pages 1321--1330. PMLR.

\bibitem[{Hendrycks et~al.(2021)Hendrycks, Burns, Kadavath, Arora, Basart, Tang, Song, and Steinhardt}]{hendrycks2021measuring}
Dan Hendrycks, Collin Burns, Saurav Kadavath, Akul Arora, Steven Basart, Eric Tang, Dawn Song, and Jacob Steinhardt. 2021.
\newblock Measuring mathematical problem solving with the math dataset.
\newblock In \emph{Thirty-fifth Conference on Neural Information Processing Systems Datasets and Benchmarks Track (Round 2)}.

\bibitem[{Hopkins et~al.(2017)Hopkins, Petrescu-Prahova, Levin, Le~Bras, Herrasti, and Joshi}]{hopkins2017beyond}
Mark Hopkins, Cristian Petrescu-Prahova, Roie Levin, Ronan Le~Bras, Alvaro Herrasti, and Vidur Joshi. 2017.
\newblock Beyond sentential semantic parsing: Tackling the math sat with a cascade of tree transducers.
\newblock In \emph{Proceedings of the 2017 Conference on Empirical Methods in Natural Language Processing}, pages 795--804.

\bibitem[{Jiang et~al.(2023)Jiang, Sablayrolles, Mensch, Bamford, Chaplot, Casas, Bressand, Lengyel, Lample, Saulnier et~al.}]{jiang2023mistral}
Albert~Q Jiang, Alexandre Sablayrolles, Arthur Mensch, Chris Bamford, Devendra~Singh Chaplot, Diego de~las Casas, Florian Bressand, Gianna Lengyel, Guillaume Lample, Lucile Saulnier, et~al. 2023.
\newblock Mistral 7b.
\newblock \emph{arXiv preprint arXiv:2310.06825}.

\bibitem[{Jiang et~al.(2024)Jiang, Sablayrolles, Roux, Mensch, Savary, Bamford, Chaplot, Casas, Hanna, Bressand et~al.}]{jiang2024mixtral}
Albert~Q Jiang, Alexandre Sablayrolles, Antoine Roux, Arthur Mensch, Blanche Savary, Chris Bamford, Devendra~Singh Chaplot, Diego de~las Casas, Emma~Bou Hanna, Florian Bressand, et~al. 2024.
\newblock Mixtral of experts.
\newblock \emph{arXiv preprint arXiv:2401.04088}.

\bibitem[{Kadavath et~al.(2022)Kadavath, Conerly, Askell, Henighan, Drain, Perez, Schiefer, Hatfield-Dodds, DasSarma, Tran-Johnson et~al.}]{kadavath2022language}
Saurav Kadavath, Tom Conerly, Amanda Askell, Tom Henighan, Dawn Drain, Ethan Perez, Nicholas Schiefer, Zac Hatfield-Dodds, Nova DasSarma, Eli Tran-Johnson, et~al. 2022.
\newblock Language models (mostly) know what they know.
\newblock \emph{arXiv preprint arXiv:2207.05221}.

\bibitem[{Kojima et~al.(2022)Kojima, Gu, Reid, Matsuo, and Iwasawa}]{kojima2022large}
Takeshi Kojima, Shixiang~Shane Gu, Machel Reid, Yutaka Matsuo, and Yusuke Iwasawa. 2022.
\newblock Large language models are zero-shot reasoners.
\newblock \emph{Advances in neural information processing systems}, 35:22199--22213.

\bibitem[{Li et~al.(2019)Li, Wang, Zhang, Wang, Dai, and Zhang}]{li2019modeling}
Jierui Li, Lei Wang, Jipeng Zhang, Yan Wang, Bing~Tian Dai, and Dongxiang Zhang. 2019.
\newblock Modeling intra-relation in math word problems with different functional multi-head attentions.
\newblock In \emph{Proceedings of the 57th annual meeting of the association for computational linguistics}, pages 6162--6167.

\bibitem[{Li et~al.(2022)Li, Zhang, Yan, Zhou, Li, Liu, and Cao}]{li2022seeking}
Zhongli Li, Wenxuan Zhang, Chao Yan, Qingyu Zhou, Chao Li, Hongzhi Liu, and Yunbo Cao. 2022.
\newblock Seeking patterns, not just memorizing procedures: Contrastive learning for solving math word problems.
\newblock In \emph{Findings of the Association for Computational Linguistics: ACL 2022}, pages 2486--2496.

\bibitem[{Magister et~al.(2023)Magister, Mallinson, Adamek, Malmi, and Severyn}]{magister2023teaching}
Lucie~Charlotte Magister, Jonathan Mallinson, Jakub Adamek, Eric Malmi, and Aliaksei Severyn. 2023.
\newblock Teaching small language models to reason.
\newblock In \emph{Proceedings of the 61st Annual Meeting of the Association for Computational Linguistics (Volume 2: Short Papers)}, pages 1773--1781.

\bibitem[{Matsuzaki et~al.(2017)Matsuzaki, Ito, Iwane, Anai, and Arai}]{matsuzaki2017semantic}
Takuya Matsuzaki, Takumi Ito, Hidenao Iwane, Hirokazu Anai, and Noriko~H Arai. 2017.
\newblock Semantic parsing of pre-university math problems.
\newblock In \emph{Proceedings of the 55th Annual Meeting of the Association for Computational Linguistics (Volume 1: Long Papers)}, pages 2131--2141.

\bibitem[{Patel et~al.(2021)Patel, Bhattamishra, and Goyal}]{patel2021nlp}
Arkil Patel, Satwik Bhattamishra, and Navin Goyal. 2021.
\newblock Are nlp models really able to solve simple math word problems?
\newblock In \emph{Proceedings of the 2021 Conference of the North American Chapter of the Association for Computational Linguistics: Human Language Technologies}, pages 2080--2094.

\bibitem[{Qin et~al.(2021)Qin, Liang, Hong, Tang, and Lin}]{qin2021neural}
Jinghui Qin, Xiaodan Liang, Yining Hong, Jianheng Tang, and Liang Lin. 2021.
\newblock Neural-symbolic solver for math word problems with auxiliary tasks.
\newblock In \emph{Proceedings of the 59th Annual Meeting of the Association for Computational Linguistics and the 11th International Joint Conference on Natural Language Processing (Volume 1: Long Papers)}, pages 5870--5881.

\bibitem[{Raffel et~al.(2020)Raffel, Shazeer, Roberts, Lee, Narang, Matena, Zhou, Li, and Liu}]{raffel2020exploring}
Colin Raffel, Noam Shazeer, Adam Roberts, Katherine Lee, Sharan Narang, Michael Matena, Yanqi Zhou, Wei Li, and Peter~J Liu. 2020.
\newblock Exploring the limits of transfer learning with a unified text-to-text transformer.
\newblock \emph{Journal of machine learning research}, 21(140):1--67.

\bibitem[{Shridhar et~al.(2023)Shridhar, Stolfo, and Sachan}]{shridhar2023distilling}
Kumar Shridhar, Alessandro Stolfo, and Mrinmaya Sachan. 2023.
\newblock Distilling reasoning capabilities into smaller language models.
\newblock In \emph{Findings of the Association for Computational Linguistics: ACL 2023}, pages 7059--7073.

\bibitem[{Touvron et~al.(2023)Touvron, Martin, Stone, Albert, Almahairi, Babaei, Bashlykov, Batra, Bhargava, Bhosale et~al.}]{touvron2023llama}
Hugo Touvron, Louis Martin, Kevin Stone, Peter Albert, Amjad Almahairi, Yasmine Babaei, Nikolay Bashlykov, Soumya Batra, Prajjwal Bhargava, Shruti Bhosale, et~al. 2023.
\newblock Llama 2: Open foundation and fine-tuned chat models.
\newblock \emph{arXiv preprint arXiv:2307.09288}.

\bibitem[{Wang et~al.(2022)Wang, Wei, Schuurmans, Le, Chi, Narang, Chowdhery, and Zhou}]{wang2022self}
Xuezhi Wang, Jason Wei, Dale Schuurmans, Quoc~V Le, Ed~H Chi, Sharan Narang, Aakanksha Chowdhery, and Denny Zhou. 2022.
\newblock Self-consistency improves chain of thought reasoning in language models.
\newblock In \emph{The Eleventh International Conference on Learning Representations}.

\bibitem[{Wei et~al.(2022)Wei, Wang, Schuurmans, Bosma, Xia, Chi, Le, Zhou et~al.}]{wei2022chain}
Jason Wei, Xuezhi Wang, Dale Schuurmans, Maarten Bosma, Fei Xia, Ed~Chi, Quoc~V Le, Denny Zhou, et~al. 2022.
\newblock Chain-of-thought prompting elicits reasoning in large language models.
\newblock \emph{Advances in neural information processing systems}, 35:24824--24837.

\bibitem[{Wu et~al.(2021)Wu, Zhang, Wei, and Huang}]{wu2021math}
Qinzhuo Wu, Qi~Zhang, Zhongyu Wei, and Xuan-Jing Huang. 2021.
\newblock Math word problem solving with explicit numerical values.
\newblock In \emph{Proceedings of the 59th Annual Meeting of the Association for Computational Linguistics and the 11th International Joint Conference on Natural Language Processing (Volume 1: Long Papers)}, pages 5859--5869.

\bibitem[{Xiong et~al.(2023)Xiong, Hu, Lu, LI, Fu, He, and Hooi}]{xiong2023can}
Miao Xiong, Zhiyuan Hu, Xinyang Lu, YIFEI LI, Jie Fu, Junxian He, and Bryan Hooi. 2023.
\newblock Can llms express their uncertainty? an empirical evaluation of confidence elicitation in llms.
\newblock In \emph{The Twelfth International Conference on Learning Representations}.

\end{thebibliography}
\bibliographystyle{acl_natbib}

\appendix



\end{document}